# Bridging National and International Legal Data: Two Projects Based on the Japanese Legal Standard XML Schema for Comparative Law Studies

Makoto Nakamura[*]

## Abstract

This paper presents an integrated framework for computational comparative law by connecting two consecutive research projects based on the Japanese Legal Standard (JLS) XML schema. The first project establishes structural interoperability by developing a conversion pipeline from JLS to the Akoma Ntoso (AKN) standard, enabling Japanese statutes to be integrated into international LegalDocML-based legislative databases. Building on this foundation, the second project applies multilingual embedding models and semantic textual similarity techniques to identify corresponding provisions across national legal systems. A prototype system combining multilingual embeddings, FAISS retrieval, and Cross-Encoder reranking generates candidate correspondences and visualizes them as cross-jurisdictional networks for exploratory comparative analysis.

**Keywords:** XML Schema, Akoma Ntoso, Legal Informatics, Comparative Law, Semantic Similarity, Legal Databases.

## CONTENTS

[*] Professor, Faculty of Engineering, Niigata Institute of Technology, Japan. The author can be contacted at mnakamur@nagoya-u.jp . This work was supported by JSPS KAKENHI Grant Numbers JP19H04427 and JP24K03230.

# I. Introduction

Understanding legal differences among countries is essential for modern societies that increasingly operate within international frameworks. Laws governing trade, environment, and technology are no longer confined to national boundaries, and legislators often refer to foreign laws when designing or revising domestic provisions. In this environment, *comparative law* serves as a key discipline that reveals how legal systems diverge, converge, and influence one another.

However, traditional comparative legal research faces enduring difficulties that arise from linguistic, cultural, and structural diversity across legal systems. The diversity of languages, legal terminologies, and historical backgrounds limits the direct comparability of legal texts. Each legal system reflects unique social values, making its interpretation highly dependent on context. Consequently, comparative studies have long relied on the judgment of experts proficient in multiple jurisdictions. Table 1 summarizes the primary aspects of this complexity.

**Table 1 Aspects of Complexity in Comparative Legal Research**

| Aspect | Explanation |
| --- | --- |
| **Language Differences** | Legal systems use different languages and legal terminology, requiring accurate translation of legal concepts and terms. |
| **Cultural and Historical Backgrounds** | Laws are shaped by the culture and history of each country, which affects how legal rules are interpreted. |
| **Systematic Structure** | Legal systems differ across countries, broadly including civil law, common law, and other traditions such as Islamic law. |
| **Document Complexity** | Legal texts often contain specialized and complex expressions that require expertise to interpret correctly. |
| **Changes in Law** | Laws are periodically amended, requiring continuous tracking of updates when comparing legal provisions. |
| **Dependence on Expertise** | Comparative analysis traditionally relies on a few multilingual experts. |

Recent advances in information technology have changed this situation. The global trend toward open government data has made legislative texts available in machine-readable formats. Japan's *e-LAWS* platform, for instance, provides all laws and regulations encoded in XML following the *Japanese Legal Standard (JLS)* schema. Meanwhile, the international community has adopted *Akoma Ntoso (AKN)* [1] as the de facto standard for representing legal documents. At the same time, natural language processing (NLP) techniques, particularly large language models based on deep learning such as BERT [2], have enabled semantic comparison of textual data across languages.

These developments allow comparative law to shift from a qualitative discipline relying on expert knowledge to a quantitative, data-driven science. Against this background, two research projects were conducted. The first project (2019–2022) focused on schema interoperability between JLS and AKN, providing technical connectivity between Japanese statutes and international legal databases [3]. The second project (2024–2028) aims to construct a multilingual database that identifies corresponding provisions among national legal systems using NLP-based semantic similarity models [4]. By combining these two achievements, this research presents a new infrastructure for *computational comparative law* that unifies the structural and semantic levels of legal information.

---

[1] Monica Palmirani and Fabio Vitali, "Akoma-Ntoso for Legal Documents," in *Legislative XML for the Semantic Web*, vol. 4, ed. Giovanni Sartor et al., Law, Governance and Technology Series (Springer Netherlands, 2011), https://doi.org/10.1007/978-94-007-1887-6_6.

[2] Jacob Devlin et al., "Bert: Pre-Training of Deep Bidirectional Transformers for Language Understanding," *Proceedings of the 2019 Conference of the North American Chapter of the Association for Computational Linguistics: Human Language Technologies, Volume 1 (Long and Short Papers)*, 2019, 4171–86, https://aclanthology.org/N19-1423/.

[3] Makoto Nakamura, "Development of Applications for Open Data for Japanese Laws and Regulations," *J. Open Access L.* 10 (2022): 1.

Beyond enabling technical interoperability, this research also seeks to contribute to the methodological transformation of comparative law itself. In recent years, the notion of *computational comparative law* has begun to emerge as a research direction that applies data-driven, reproducible, and scalable methods to cross-jurisdictional legal analysis[5], reflecting the broader integration of computational methods into legal scholarship. Unlike traditional doctrinal comparison, which relies heavily on expert interpretation of selected provisions, computational approaches attempt to process entire legislative corpora and identify structural or semantic patterns at scale.

The present study does not claim to replace doctrinal expertise. Rather, it aims to construct an infrastructure that augments expert analysis by generating correspondence candidates, visualizing structural proximity, and enabling systematic exploration of large bodies of legal text. At its current stage, the system remains a prototype and does not yet provide fully validated gold-standard evaluation results. Instead, it demonstrates the feasibility of integrating structural standardization and multilingual semantic modeling into a unified comparative-law platform.

This paper is organized as follows. Section II reviews related work on legal data standardization, reference-network analysis in comparative law, and recent developments in legal informatics and multilingual NLP. Section III presents the overall research framework that integrates structural interoperability and semantic mapping into a unified pipeline. Section IV describes Project 1, which establishes schema interoperability through the conversion of Japanese Legal Standard (JLS) XML into the Akoma Ntoso (AKN) format, thereby enabling international structural connectivity. Section V introduces Project 2 and explains the architecture of the multilingual comparative-law database based on BERT and semantic textual similarity (STS) modeling. Section VI presents the construction of the cross-national corresponding-article extraction system and reports the current progress of the prototype implementation, including embedding-based retrieval, reranking with a Cross-Encoder model, and graph-based visualization of correspondences. Section VII discusses the integration of structural and semantic layers and reflects on the methodological implications for computational comparative law. Finally, Section VIII concludes the paper by summarizing the contributions of both projects and outlining future research directions toward a globally interoperable legal information infrastructure.

---

[4] Hiroki Cho et al., "Mapping Similar Provisions Between Japanese and Foreign Laws," in *New Frontiers in Artificial Intelligence*, vol. 13859, ed. Yasufumi Takama et al., Lecture Notes in Computer Science (Springer Nature Switzerland, 2023), https://doi.org/10.1007/978-3-031-29168-5_3.

[5] Jaakko Husa, "Comparative Law's Pyrrhic Victory?," *Maastricht Journal of European and Comparative Law* 30, no. 6 (2023): 680–88, https://doi.org/10.1177/1023263X241252517; Jason King, "Computational Equivalence: A Structured Lab Methodology for Comparative Law in the Age of Artificial Intelligence  ," *SSRN Electronic Journal*, ahead of print, 2026, https://doi.org/10.2139/ssrn.5908502.

## II. Related Work

### A Legal Data Standardization and Open Access

Since the 1990s, the Free Access to Law Movement (FALM) has promoted public access to legal information. Initiatives such as LII, AustLII, and SAFLII provided online databases of statutes and case law. While the Japan Legal Information Institute (JaLII) was not an official FALM member, its contributions paralleled the global open-law movement and provided a domestic model for structured legislative data. JaLII developed the Japanese Legal Standard (JLS) schema, which later became the foundation of the government's e-LAWS system. Earlier versions of the JLS were created using Document Type Definitions (DTD) and were employed in the Japan Law Translation Database System[6][7].

Subsequently, the Akoma Ntoso (AKN) schema [8] —standardized under LegalDocML by OASIS—emerged as an international framework for legislative representation. AKN introduces a consistent hierarchical structure (*act*, *part*, *chapter*, *article*) and metadata elements, which define the identity and version of each law. Through this framework, legal documents from different jurisdictions can be uniformly parsed and linked.

Japan's JLS and AKN share many structural similarities but differ in tag naming, metadata granularity, and namespace conventions. The conversion of JLS to AKN enables interoperability between Japan's domestic legal corpus and other AKN-based international legislative databases. This structural connection laid the foundation for Project 1 introduced in this paper.

---

[6] https://www.japaneselawtranslation.go.jp

[7] Katsuhiko Toyama et al., "Design and Development of the Japanese Law Translation Database System," *Law via the Internet Conference Proceedings*, 2011; Katsuhiko Toyama et al., "Design and Development of the Japanese Law Translation Database System (in Japanese)," *Journal of Information Network Law Review* 11 (2012): 33–53.

[8] Palmirani and Vitali, "Akoma-Ntoso for Legal Documents."

## B Reference Networks and Comparative Law

In comparative law, quantitative approaches to analyzing the structural characteristics of civil codes have recently attracted increasing attention. A representative example is the study by Badawi and Dari-Mattiacci [9], which compared the network structures of cross-references among the *Code civil* (France), the *Bürgerliches Gesetzbuch* (BGB, Germany), and the *Allgemeines Bürgerliches Gesetzbuch* (ABGB, Austria).

In their analysis, each article of a code is modeled as a node and each reference between articles as an edge, enabling measurement of network density, hierarchy, and modularity. The results supported the conventional view that the German Civil Code was designed to employ cross-references more extensively than the French Civil Code. Furthermore, although the Austrian Civil Code was enacted around the same period (early 1800s), it exerted only limited influence on other jurisdictions, whereas the French and German codes strongly shaped subsequent codifications abroad. These findings quantitatively clarified the internal organization and international diffusion of civil-code structures, marking an important step toward a structural and quantitative approach to comparative law.

This line of research goes beyond traditional interpretive comparisons based solely on wording or doctrinal concepts, offering a framework for objectively capturing how legal systems are organized and interrelated.

Figure 1 illustrates this concept. Building on the reference-network analysis of Badawi and Dari-Mattiacci [10], it adds the Japanese Civil Code to the network of the French, German, and Austrian codes, and connects functionally corresponding provisions across codes with polylines representing "skewering" correspondences; that is, cross-cutting links across jurisdictions. The figure serves only as a conceptual example to visualize the idea of cross-jurisdictional linkage; it does not represent actual mapping data. Through this illustrative extension, Figure 1 shows how such skewering correspondences might span multiple legal systems.

---

[9] Adam B. Badawi and Giuseppe Dari-Mattiacci, "Reference Networks and Civil Codes," *See Livermore & Rockmore* 2019 (2019): 339–65.

[10] Badawi and Dari-Mattiacci, "Reference Networks and Civil Codes."

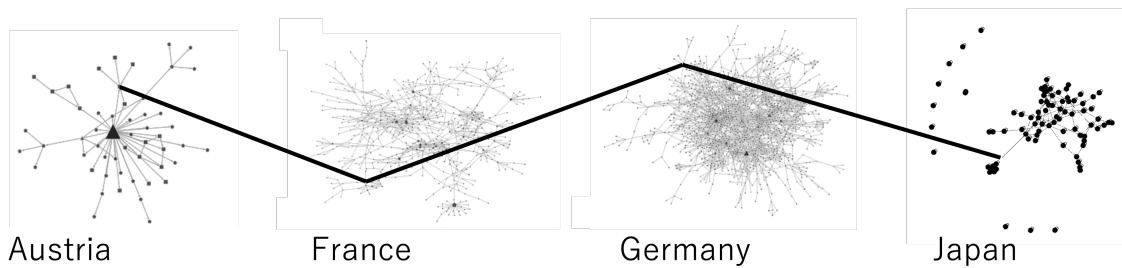

| Austria | France | Germany | Japan |

**Figure 1 Example of Inter-Code Correspondences Based on Badawi and Dari-Mattiacci** [11]
*This figure extends the reference network among the French, German, and Austrian Civil Codes by adding the Japanese Civil Code. Functional "skewering" correspondences between semantically similar provisions are illustrated with polylines. This figure is provided solely for conceptual explanation and does not represent actual mapping data.*

### C Natural Language Processing and Legal Informatics

In parallel with the standardization of legal documents, research in legal informatics has advanced rapidly. Early comparative analyses relied on lexical similarity measures such as TF–IDF, which were insufficient when different jurisdictions expressed similar ideas in distinct wording. With the introduction of deep-learning-based language models such as BERT [12] and its multilingual variants, it became possible to evaluate *semantic textual similarity (STS)* directly. Legal-domain adaptations such as *LegalBERT* [13] have further improved performance in understanding statutory language.

Recent initiatives such as the *Competition on Legal Information Extraction and Entailment (COLIEE)* [14] have demonstrated the potential of NLP in legal reasoning. COLIEE includes tasks drawn from both civil-law and common-law jurisdictions and focuses primarily on answering bar-exam-style legal questions that require reasoning and entailment across case law and statutory texts. Although this framework has contributed significantly to benchmarking legal AI systems, its objective is to test reasoning capability rather than to analyze structural relationships among legal systems.

In contrast, the present research aims to quantify comparative law by establishing correspondences among statutes across multiple jurisdictions. Instead of solving legal questions, our goal is to measure semantic relationships between legal provisions and visualize how legal concepts propagate across national boundaries. Thus, while COLIEE represents an applied, competition-oriented approach to legal AI, this study positions itself as a foundational contribution to data-driven comparative jurisprudence.

---

[11] Badawi and Dari-Mattiacci, "Reference Networks and Civil Codes."

[12] Devlin et al., "Bert."

[13] Ilias Chalkidis et al., "LEGAL-BERT: The Muppets Straight out of Law School," *Findings of the Association for Computational Linguistics: EMNLP 2020*, 2020, 2898–904, https://aclanthology.org/2020.findings-emnlp.261/.

[14] Juliano Rabelo et al., "Overview and Discussion of the Competition on Legal Information Extraction/Entailment (COLIEE) 2021," *The Review of Socionetwork Strategies* 16, no. 1 (2022): 111–33, https://doi.org/10.1007/s12626-022-00105-z.

While previous work has addressed either structural representation (e.g., XML schema standardization) or semantic modeling (e.g., legal-domain BERT models), relatively few studies have integrated both dimensions into a single coherent pipeline. Structural interoperability alone does not guarantee semantic comparability, and semantic similarity without standardized document structure may lead to unstable or context-insensitive alignments. Computational comparative law therefore requires a dual foundation: formal structural normalization and cross-lingual semantic modeling. The present research explicitly situates itself at this intersection, proposing an integrated framework in which schema conversion and semantic similarity analysis mutually reinforce one another.

## III. Research Framework

Section II reviewed prior research on structural standardization and semantic modeling in legal informatics, which have often developed as parallel but largely independent strands. The present section integrates these strands into a unified framework for computational comparative law. The proposed architecture combines schema-level interoperability and cross-lingual semantic similarity within a continuous pipeline, transforming structured legislative corpora into graph-based representations of inter-jurisdictional relationships.

The two projects described in this paper jointly implement this integrated framework. The first project ensures structural compatibility by converting Japanese statutes from JLS to AKN, while the second identifies semantic correspondences between AKN-encoded laws of different countries. Together, they create a data-driven infrastructure for large-scale comparative analysis.

Figure 2 illustrates the overall framework of this research. The Input Layer consists of legislative corpora in JLS and AKN formats. Project 1 implements the Schema Interoperability Layer through schema conversion (JLS→AKN), while Project 2 provides the Semantic Mapping Layer based on multilingual BERT and STS. These outputs are combined in the Integration & Visualization Layer (Output) to produce a Global Comparative Law Graph that connects provisions across jurisdictions.

Through these four layers, legal texts from Japan and other jurisdictions can be compared automatically, both structurally and semantically.

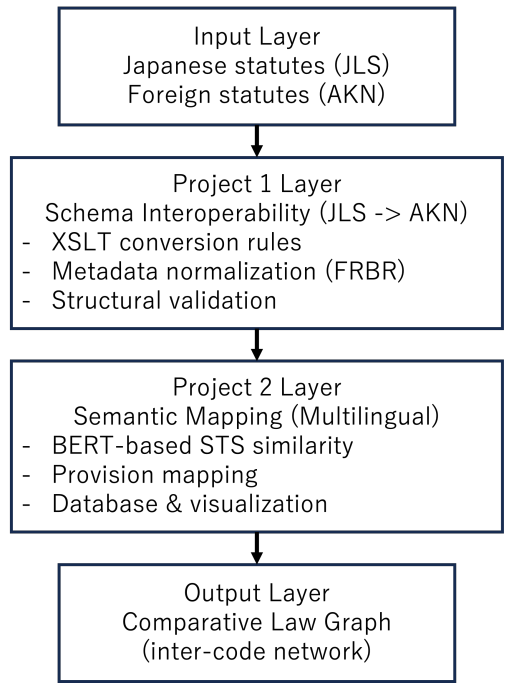

**Figure 2 Overall Framework of the Research Projects**
*Input → Schema Interoperability (JLS→AKN) → Semantic Mapping (BERT/STS) → Integration & Visualization (Output). The final layer consolidates structural and semantic links into the Global Comparative Law Graph.*

## IV.  Project 1 – Schema Conversion and International Connectivity

### A  Objectives and Background

The first project, *Development of Major Applications for Making Effective Use of Japanese Statutes as Open Data* (2019–2022), was undertaken to enhance interoperability between Japan's domestic legislative corpus and international legal information standards [15]. While the *e-LAWS* system successfully standardized Japanese statutes using the *Japanese Legal Standard (JLS)* schema, the resulting structure was incompatible with the *Akoma Ntoso (AKN)* schema widely used in Europe and other jurisdictions. This limitation prevented Japanese legal data from being incorporated into cross-national research and multilingual databases. The project therefore developed a complete conversion pipeline from JLS to AKN, ensuring both structural and semantic alignment with international legislative corpora. It should be noted that compatibility with the AKN schema had already been confirmed in the earlier DTD-based version of this initiative [16].

---

[15] Nakamura, "Development of Applications for Open Data for Japanese Laws and Regulations."

[16] Gen Kawachi et al., "Applying the Akoma Ntoso XML Schema to Japanese Legislation," *Journal of Law, Information and Science* 24, no. 2 (2016): 24–102.

## B  Conversion Pipeline and Methodology

The schema-conversion pipeline developed in this project consists of six stages:

1. **Input and Parsing:** JLS XML files are collected from *e-LAWS* and parsed to extract hierarchical elements such as <Law>, <Article>, and <Paragraph>.
2. **Structural Analysis:** Each element is examined to determine its logical level within the legislative hierarchy and to resolve inconsistencies prior to conversion.
3. **XSLT Transformation:** About 50 mapping rules, defined in the converter, transform JLS elements into their AKN equivalents (e.g., <Law> → <act>, <Article> → <article>).
4. **Metadata Generation (FRBR Model):** Identifiers for *Work*, *Expression*, and *Manifestation* are assigned, and additional information such as jurisdiction, language, and version date is embedded.
5. **Validation and Verification:** The converted files are validated against the LegalDocML AKN schema and checked for structural integrity and namespace consistency. Ten randomly selected laws were manually verified, and all passed schema validation.
6. **Visualization and Editing:** The validated AKN files can be displayed and edited using open-source applications such as *LIME Editor* and *Akomantoso Viewer*.

This process ensures structural interoperability between Japanese legislative data and international AKN-based systems, thereby establishing a foundation for cross-jurisdictional comparative analysis.

## C  Hierarchical Mapping and Connectivity

The conversion is grounded in a one-to-one hierarchical correspondence between JLS and AKN elements, as illustrated in Figure 3. Each level of the Japanese schema, from <Law> to <Article> and <Paragraph>, is mapped to its AKN equivalent (act, article, paragraph respectively). This structural alignment allows Japan's laws to be integrated into international databases without loss of logical integrity.

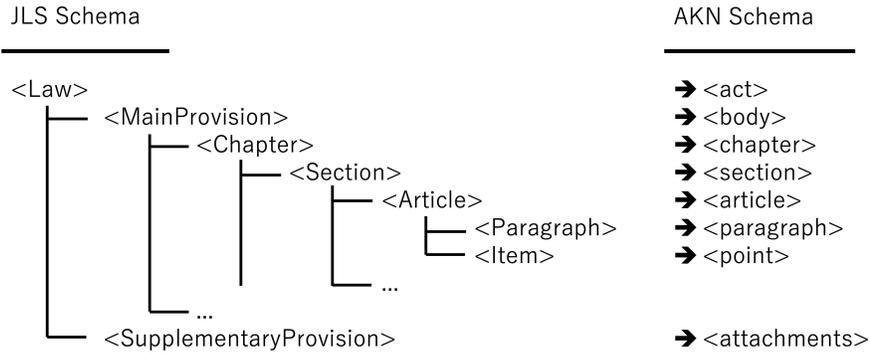

**Figure 3 Hierarchical Mapping between JLS and AKN Schema**
*Each box represents a hierarchical level in the legislative document structure. Arrows indicate the element correspondence between the Japanese Legal Standard (JLS) and the Akoma Ntoso (AKN) schema.*

### D  Validation and Outcomes

Through the conversion pipeline developed in Project 1, Japanese legislative data can now be parsed, validated, and displayed within any Akoma Ntoso-compliant environment. This confirms that the structural design of the Japanese Legal Standard (JLS) schema is technically compatible with the AKN framework used in international legislative systems. The validation process ensured conformity with the LegalDocML-AKN schema and preserved the logical integrity of every converted law.

Accordingly, the outcome of Project 1 goes beyond simple schema engineering: it establishes a technical foundation that connects Japan's legislative corpus to international open standards and supports the semantic correspondence analysis developed in Project 2.

Although the validation experiment was primarily technical in nature, focusing on schema conformity and structural integrity, its implications extend beyond engineering compatibility. The successful transformation of representative statutes indicates that the hierarchical and conceptual design embedded in the Japanese Legal Standard (JLS) schema is fundamentally compatible with the LegalDocML-AKN framework. This compatibility suggests that national legislative drafting structures can be harmonized at the level of formal representation without erasing jurisdiction-specific characteristics. In this sense, Project 1 provides not only structural interoperability but also a formal bridge between domestic legislative identity and international legal-data standards.

# V.　　Project 2 – Comprehensive Comparative Law Database

## A　Objectives and Background

Building upon the structural interoperability achieved in Project 1, Project 2 aims to construct a comprehensive multilingual comparative-law database that enables the automatic identification of corresponding provisions among different jurisdictions. The goal is to advance comparative law from a qualitative, expert-driven discipline toward a more quantitative and reproducible approach based on natural-language processing and standardized legislative data.

Traditional comparative-law studies have relied heavily on manual interpretation. While such approaches offer deep contextual understanding, they lack scalability and reproducibility. By contrast, Project 2 integrates linguistic modeling, database engineering, and visualization technologies to enable large-scale cross-jurisdictional comparison. This approach enhances the precision of semantic alignment and provides a unified research infrastructure for comparative jurisprudence.

By integrating standardized AKN-encoded corpora with multilingual semantic embeddings, Project 2 extends structural interoperability into the semantic domain. This integration transforms legislative corpora from static repositories into analyzable semantic networks, enabling quantitative comparison beyond textual surface similarity.

## B　System Architecture

The overall workflow of the system is illustrated in Figure 4, consisting of six modules labeled [a]–[f]. Each module represents a distinct stage in the process of constructing and utilizing a multilingual comparative-law platform.

**[a]** The **World Law Database** stores legislative texts from multiple jurisdictions, including Japan. These corpora serve as the foundational data for cross-jurisdictional analysis.

**[b]** From these legislative datasets, a **World-Legal BERT** model is created through pre-training. The model learns multilingual and cross-domain representations that capture semantic relations among legal provisions.

**[c]** Based on the pre-trained model, the system computes **Semantic Textual Similarity (STS)** scores using law data from each jurisdiction. This enables semantic comparison of provisions across different legal systems.

**[d]** As a result, the similarity between articles is calculated, and each Japanese provision is **linked to the corresponding article of foreign law** with the highest similarity score. This creates a one-to-many network of semantically related provisions.

**[e]** In response to a user's query, the system retrieves and displays the relevant provisions from multiple jurisdictions, together with their associated similarity information.

**[f]** Finally, the **visualized data,** including semantic links and correspondence graphs, are generated as output, allowing users to intuitively explore relationships among legal systems.

Through this architecture, legal texts from Japan and other jurisdictions can be semantically analyzed, compared, and visualized within a unified environment.

For example, a user may ask a question through a large language model (LLM) interface [e] such as: "How do fathers of unmarried children differ around the world?" In response to this query, the system searches the World Law Database [a] to identify relevant legal provisions. The retrieved provisions are then linked with semantically related foreign statutes that have been pre-associated through the correspondence extraction process, and the resulting cross-jurisdictional relationships are presented through the visualization module [f]. Through this mechanism, the system enables a question-answering environment in which users can explore how similar legal concepts are treated across different legal systems.

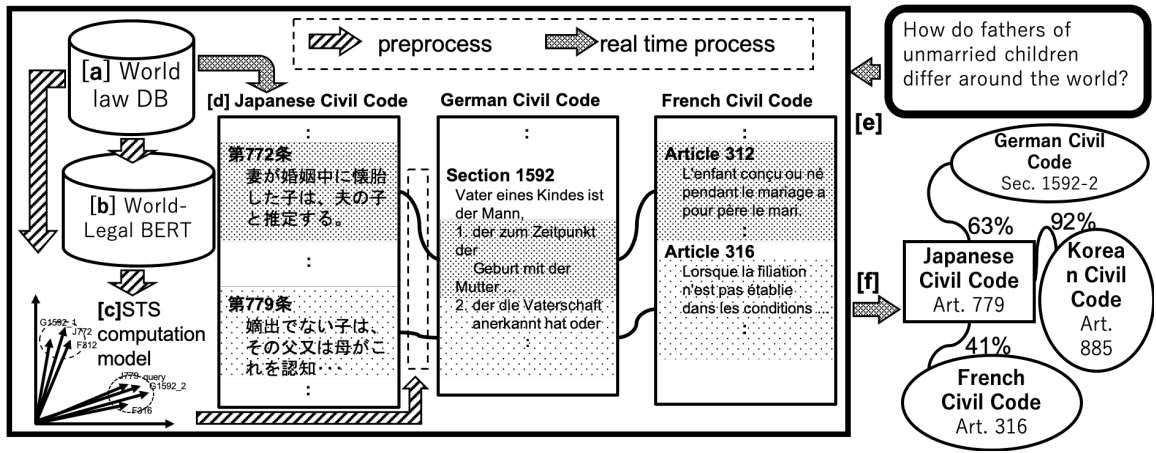

**Figure 4 System Architecture of Project 2:** *[a] World Law Database   [b] Pre-training of World-Legal BERT   [c] STS Computation   [d] Article Linking   [e] Query and Retrieval   [f] Visualization and Output*

### C  Previous Experiments on Legal Provision Mapping

Experiments in Project 2 investigated semantic correspondence detection between legal provisions using several similarity-based approaches. The methods examined ranged from word-based models such as TF-IDF and BM25 to deep-learning models including Sentence-BERT and fine-tuned variants. The evaluation datasets included both intra-language correspondence tasks between Japanese statutes and cross-jurisdictional mappings between the Japanese Civil Code and the German Civil Code, using original texts as well as available English and Japanese translations.

To evaluate the effectiveness of the proposed approach, gold-standard correspondence datasets were prepared for two types of experiments: a domestic-law mapping task and a cross-jurisdictional mapping task.

For the domestic experiment, we constructed a dataset aligning provisions between the Japanese Electricity Business Act and Gas Business Act. Because many provisions in these two statutes are structurally and functionally similar, corresponding articles were manually identified to create the gold-standard dataset.

For the cross-jurisdictional experiment, we prepared a dataset for the family-law sections of the Japanese Civil Code and the German Civil Code. The correspondence pairs were derived from the "comparison" sections provided in the article-by-article commentary on the Civil Code, which explicitly indicate related provisions in foreign legal systems.

Using these datasets, we evaluated the semantic textual similarity (STS) approach. For the domestic mapping between the Electricity Business Act and the Gas Business Act, the system achieved an F1 score of 0.768 [17]. For the cross-jurisdictional mapping between the Japanese and German Civil Codes, the F1 score was 0.348 [18].

These results indicate that semantic mapping can identify corresponding provisions even across different legal systems and languages. However, the lower score in the cross-jurisdictional experiment reflects the additional complexity introduced by linguistic differences, translation variation, and conceptual divergence between legal traditions.

## VI. A Prototype System for Cross-National Correspondence Extraction

This section presents the current prototype implementation designed to explore automatic extraction of cross-national corresponding provisions.

### A Objectives and Target Corpora

The objective of this prototype is to construct a structurally standardized and semantically analyzable corpus that enables exploratory comparative analysis across multiple jurisdictions.

Legislative corpora were prepared in the Akoma Ntoso XML format for the Civil Codes and Commercial Codes of Japan, Korea, France, and Germany. Although these corpora do not yet constitute fully validated and complete digital editions, the structural encoding enables the specification and extraction of text at multiple hierarchical levels, including articles, paragraphs, and items. This hierarchical representation preserves formal legislative organization while providing text data suitable for computational processing.

---

[17] Cho et al., "Mapping Similar Provisions Between Japanese and Foreign Laws."

[18] Toshinori Takahashi et al., "A Multilingual Legal Provision Mapping Across Jurisdictions: A One-to-Many Approach," in *New Frontiers in Artificial Intelligence*, vol. 15692, ed. Yukiko Nakano and Toyotaro Suzumura, Lecture Notes in Computer Science (Springer Nature Singapore, 2025), https://doi.org/10.1007/978-981-96-7071-0_10.

For the experiment reported in this paper, the analysis focuses on the Civil Code datasets of Japan, Korea, and France. By restricting the scope to these three jurisdictions, the prototype examines cross-national correspondences within a comparable doctrinal domain. The use of the Akoma Ntoso format ensures that structural elements such as article numbering and hierarchical organization are retained while enabling semantic modeling and cross-lingual similarity computation.

### B  Semantic Representation and Candidate Retrieval

To identify potentially corresponding provisions across jurisdictions, each article is first converted into a vector representation using the multilingual embedding model multilingual-e5-large[19]. This model is based on the Transformer architecture[20] and a RoBERTa-based pretraining framework[21], produces multilingual semantic embeddings optimized for retrieval tasks. By representing each article as a dense vector in a shared embedding space, provisions with similar legal meaning can be located close to one another even when written in different languages.

Because the number of provisions across the target legal corpora reaches several thousand, exhaustive pairwise comparison is computationally impractical. Instead, we employ FAISS[22] (Facebook AI Similarity Search) to perform efficient approximate nearest-neighbor search. In this stage, each article of the Japanese Civil Code is used as a query, and candidate articles are retrieved from the Korean and French Civil Codes based on vector similarity. The objective of this stage is to collect a sufficiently broad set of candidate correspondences while maintaining computational efficiency.

For each Japanese article, the system first retrieves up to 120 candidate articles from the vector index (topk_cand = 120). From this candidate pool, the system retains up to 80 candidates per country (pre_per_country = 80), restricting the target jurisdictions to Korea and France. This country-aware candidate selection step ensures that subsequent analysis focuses on cross-jurisdictional correspondences rather than intra-system similarities.

---

[19] Liang Wang et al., "Text Embeddings by Weakly-Supervised Contrastive Pre-Training," arXiv:2212.03533, preprint, arXiv, February 22, 2024, https://doi.org/10.48550/arXiv.2212.03533.

[20] Ashish Vaswani et al., "Attention Is All You Need," *Advances in Neural Information Processing Systems* 30 (2017), https://papers.nips.cc/paper_files/paper/2017/hash/3f5ee243547dee91fbd053c1c4a845aa-Abstract.html.

[21] Yinhan Liu et al., "RoBERTa: A Robustly Optimized BERT Pretraining Approach," arXiv:1907.11692, preprint, arXiv, July 26, 2019, https://doi.org/10.48550/arXiv.1907.11692.

[22] Jeff Johnson et al., "Billion-Scale Similarity Search with GPUs," *IEEE Transactions on Big Data* 7, no. 3 (2019): 535–47.

### C  Multi-Stage Reranking and Network Construction

The extraction of corresponding provisions is performed through a multi-stage retrieval and reranking process designed to balance semantic relevance and cross-jurisdictional comparability.

First, for each Japanese provision, the embedding-based nearest-neighbor search retrieves a large candidate set. From this set, the system keeps up to 80 candidates per country and then applies a country-level quota so that the number of candidates from each jurisdiction remains balanced. Specifically, the number of retained candidates is limited to 30 per country (quota_per_country = KR = 30, FR = 30). This constraint prevents candidate dominance by a single jurisdiction and ensures that the comparison remains structurally balanced across legal systems.

Next, the remaining candidate pairs are evaluated using a Cross-Encoder reranking model[23], which jointly encodes each pair of provisions and directly estimates their semantic correspondence. Compared with embedding-based similarity alone, this approach allows more precise evaluation of doctrinal consistency between provisions. After reranking, the system outputs up to 60 candidate correspondences per Japanese article (topk_final = 60), which are stored in a JSON file as structured correspondence candidates.

The generated correspondences are subsequently visualized as a bipartite network representation. Figure 5 shows the resulting network of correspondences between Japanese, Korean, and French legal provisions obtained under the country-specific threshold settings. In this graph, Japanese provisions are placed in the central column, while Korean and French provisions are arranged in the left and right columns, respectively. This column-based layout enables intuitive inspection of cross-jurisdictional relationships centered on Japanese law.

---

[23] Nils Reimers and Iryna Gurevych, "Sentence-Bert: Sentence Embeddings Using Siamese Bert-Networks," *Proceedings of the 2019 Conference on Empirical Methods in Natural Language Processing and the 9th International Joint Conference on Natural Language Processing (EMNLP-IJCNLP)*, 2019, 3982–92, https://aclanthology.org/D19-1410/.

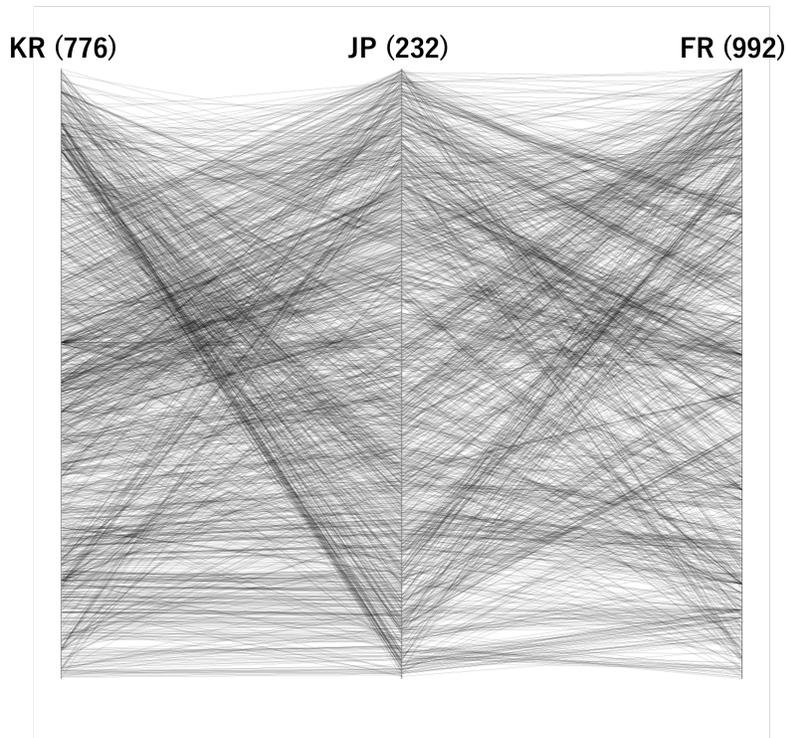

Figure 5. Correspondence network between Korean, Japanese, and French Civil Code provisions generated by the proposed prototype system. The graph contains 2,000 nodes (KR: 776, JP: 232, FR: 992) and 3,545 edges. Edges represent candidate correspondences identified through multilingual embeddings, FAISS retrieval, and Cross-Encoder reranking.

During visualization, country-specific similarity thresholds and edge limits are applied in order to maintain comparable correspondence density across jurisdictions. For each Japanese article, at most three Korean correspondences with similarity scores of 0.95 or higher are retained, while for French provisions at most three correspondences with similarity scores of 0.80 or higher are accepted as edges. By controlling both similarity thresholds and edge counts separately for each country, the system adjusts the relative density of Japan–Korea and Japan–France correspondences so that the resulting network remains analytically comparable.

Under these settings, the resulting correspondence network consists of 2,000 vertices and 3,545 edges. The network provides a visual representation of semantic proximity between provisions and serves as an exploratory infrastructure for identifying potential correspondences and analyzing structural relationships among legal systems in comparative law research.

### D  Prototype Limitations and Methodological Implications

At the present stage, this research remains at the prototype development phase. The objective of this stage is therefore to demonstrate the feasibility of automatically generating and structuring candidate correspondences rather than to provide statistically validated accuracy results. The correspondences produced by the system should be regarded as exploratory candidates that require further expert evaluation. Future work will include the construction of expert-annotated gold-standard datasets, the development of evaluation metrics, and systematic verification of correspondence accuracy.

The significance of this study lies not merely in computing textual similarity between provisions, but in providing an infrastructure for cross-system visualization of normative structures across multiple legal systems. In comparative law research, article-by-article comparisons have traditionally been conducted for specific issues; however, the framework proposed here enables comprehensive generation of correspondence candidates across entire codes. This allows researchers not only to examine individual correspondences but also to analyze structural proximity and concentrations of alignment at the level of chapters or books.

Furthermore, the system serves as a tool for exploratory research. It has the potential to reveal latent correspondences between provisions that may not have been explicitly recognized in prior scholarship, thereby expanding the starting points for comparative legal discussions. In the future, by incorporating additional jurisdictions, expanding to further areas of law, and developing evaluation datasets, we aim to develop this database into foundational infrastructure for comparative legal research.

## VII. Integration and Discussion

### A  From Structure to Semantics

Project 1 achieved schema interoperability through the conversion of Japanese legislative data from JLS to AKN, while Project 2 implemented semantic interoperability based on multilingual BERT and STS modeling. By combining these two layers, the system establishes a continuous pipeline from structural encoding (JLS → AKN) to semantic correspondence analysis (BERT → STS → Database). This integration forms a unified infrastructure for computational comparative law, connecting Japan's domestic legal corpus with international standards and enabling automatic cross-jurisdictional comparison. By linking structural encoding and semantic similarity within a graph-based representation, the system implicitly models law as a multi-layered normative network. In this perspective, legislative provisions are not isolated textual units but interconnected nodes within evolving legal ecosystems. Structural interoperability ensures formal coherence, while semantic alignment captures functional proximity across jurisdictions. The integration of these layers thus reframes comparative law as the analysis of inter-systemic relational patterns rather than merely textual comparison.

### B  Methodological Implications for Computational Comparative Law

The integration of Projects 1 and 2 demonstrates that Japan's legislative data, standardized under the JLS schema, can be extended to support multilingual and multi-jurisdictional research within the AKN framework. Technically, the system bridges the gap between structural standardization and semantic alignment, providing an operational model for interoperable legislative data. Academically, it establishes a quantitative approach to comparative law, contributing to the emerging field of computational comparative law by combining legal informatics and cross-lingual semantics.

Future work will focus on improving similarity computation through refined domain-specific models, addressing linguistic and conceptual divergences across jurisdictions. In addition, the corpus will be expanded primarily to other civil law jurisdictions, enabling broader comparative coverage beyond the currently analyzed legal systems. These developments will support a more comprehensive analysis of legal evolution and inter-jurisdictional influence, thereby contributing to a deeper global understanding of legal diversity in the digital era.

### C  Academic Significance

This study contributes to the field of computational comparative law by integrating structural and semantic interoperability between legislative corpora. Academically, it establishes a quantitative framework for analyzing inter-jurisdictional correspondences, expanding the methodological scope of comparative law. Practically, the proposed infrastructure enhances the accessibility and reusability of legislative data, facilitating cross-border legal research and international collaboration. The unified schema and similarity-based mapping will also support multilingual legal information systems and promote transparency in legislative processes. At the current stage, the system functions primarily as a research prototype. Although initial experiments demonstrate the feasibility of multilingual semantic alignment, comprehensive evaluation using large-scale expert-annotated datasets remains an important future task. Establishing reproducible benchmarking protocols will be essential for ensuring methodological transparency and long-term academic reliability.

### D  Future Development

Although a full prototype interface has not yet been implemented, the system design anticipates an interactive environment in which users can query, retrieve, and compare corresponding provisions across jurisdictions. Future work will involve developing this interface, integrating LLM-based semantic search or explanation modules, and enabling comparative visualization of one-to-many semantic correspondences. These functions will make the comparative-law database more accessible to legal scholars and practitioners, bridging the gap between research and applied legal informatics.

## VIII. Conclusion

This paper has presented an integrated research program aimed at establishing a foundation for computational comparative law by combining structural interoperability and semantic alignment across national legislative corpora. Rather than treating legal texts as either formally structured documents or purely linguistic artifacts, the study has demonstrated that meaningful cross-jurisdictional comparison requires the coordinated integration of both dimensions.

Project 1 achieved structural interoperability by developing a complete and validated conversion pipeline from the Japanese Legal Standard (JLS) schema to the Akoma Ntoso (AKN) framework. This conversion enables Japanese legislative data to participate in international LegalDocML-based environments without loss of hierarchical integrity or metadata coherence. The results confirm that the structural logic embedded in the JLS schema is conceptually compatible with global legislative data standards, thereby establishing a formal bridge between domestic legislative identity and international interoperability.

Building upon the structural alignment framework established in Project 1, Project 2 aims to extend the analysis into the semantic domain in order to identify corresponding legal provisions across jurisdictions. As discussed in Project 2, earlier experiments on semantic correspondence detection showed that when both legal texts were written in Japanese, such as in correspondence detection between Japanese statutes, the proposed approaches achieved relatively high levels of accuracy. However, when the same methods were applied to cross-jurisdictional comparisons involving foreign civil codes, the accuracy tended to decrease due to linguistic differences, variations in legal terminology, and conceptual divergences across legal systems.

Section VI presented a prototype system integrating semantic retrieval, reranking, and visualization components to generate cross-national correspondence candidates between Japanese, Korean, and French legal provisions. Although the current implementation remains at a prototype stage and does not yet rely on large-scale expert-annotated gold-standard datasets, it demonstrates the feasibility of constructing an infrastructure for cross-lingual semantic alignment of legislative provisions. The resulting correspondences can be externalized as graph-based representations that visualize inter-systemic relationships among legal systems. Accordingly, the present results should be interpreted as a proof of concept demonstrating infrastructural feasibility rather than definitive measurements of doctrinal equivalence.

Taken together, the integration of schema conversion and semantic modeling transforms Japan's legislative corpus from a nationally bounded dataset into a node within a globally interoperable legal information network. More broadly, this research suggests that comparative law in the digital era can evolve from selective doctrinal juxtaposition toward large-scale, reproducible, and data-driven analysis of normative structures. By modeling law as a multi-layered network that combines formal hierarchy and semantic proximity, computational comparative law enables the systematic exploration of structural alignment, conceptual diffusion, and patterns of inter-jurisdictional influence.

Future research will focus on expanding the coverage of the system in order to achieve comprehensive article-level correspondence across jurisdictions. This includes increasing the number of statutes analyzed, designing an interactive user interface, and extending the corpus primarily to additional civil law jurisdictions. To support these developments, it will be necessary to construct expert-annotated gold-standard correspondence datasets that enable systematic evaluation and refinement of the correspondence extraction methods.

In conclusion, this study does not claim to replace traditional comparative legal scholarship. Instead, it proposes an infrastructural and methodological foundation that augments expert analysis with scalable computational tools. By unifying structural standardization and semantic modeling within a single integrated architecture, the research advances the emergence of computational comparative law as a reproducible and internationally collaborative research paradigm.